\documentclass{INTERSPEECH2023}

\interspeechcameraready 

\usepackage{color}

\newcommand{\benchmark}{\textit{Svarah}}

\title{\benchmark: Evaluating English ASR Systems on Indian Accents}
\name{Tahir Javed$^{1,2}$, Sakshi Joshi$^2$, Vignesh Nagarajan$^2$, Sai Sundaresan$^2$, Janki Nawale$^2$, Abhigyan Raman$^2$, Kaushal Bhogale$^{1,2}$, Pratyush Kumar$^{1,2,3}$, Mitesh M. Khapra$^{1,2}$}
\address{
  $^1$Indian Institute of Technology Madras, India\\
  $^2$AI4Bharat, India \\
  $^3$Microsoft, India
  }
\email{{tahirjmakhdoomi,sakshijcom,vignesh.vn.nagarajan,saisundaresan01}@gmail.com, janki@ai4bharat.org, {cs22d006,pratyush,miteshk}@cse.iitm.ac.in}

\begin{document}

\maketitle
 
\begin{abstract}

India is the second largest English-speaking country in the world with a speaker base of roughly 130 million. 
Thus, it is imperative that automatic speech recognition (ASR) systems for English should be evaluated on Indian accents.
Unfortunately, Indian speakers find a very poor representation in existing English ASR benchmarks such as LibriSpeech, Switchboard, Speech Accent Archive, etc. 
In this work, we address this gap by creating \benchmark, a benchmark that contains 9.6 hours of transcribed English audio from 117 speakers across 65 geographic locations throughout India, resulting in a diverse range of accents. 
\benchmark~comprises both read speech and spontaneous conversational data, covering various domains, such as history, culture, tourism, etc., ensuring a diverse vocabulary. 
We evaluate 6 open source ASR models and 2 commercial ASR systems on \benchmark~and show that there is clear scope for improvement on Indian accents. 
\benchmark~as well as all our code will be publicly available.

\end{abstract}

\section{Introduction}
Recent advances in Automatic Speech Recognition have demonstrated that current models have achieved human parity in transcription \cite{whisper, 8049322}. For example, state-of-the-art systems from Google and Microsoft have reported Word Error Rates (WERs) as low as 4.9\%\footnote{https://venturebeat.com/business/googles-speech-recognition-technology-now-has-a-4-9-word-error-rate/} and 5.1\% \cite{8049322} respectively. Similarly, open-source models such as Meta's Wav2vec2 \cite{wav2vec2} and OpenAI's Whisper \cite{whisper} have reported WERs of 1.8\% and 2.7\% on the LibriSpeech benchmark. However, these benchmarks only contain data from native speakers and a more robust evaluation using benchmarks with a diverse representation of accents is missing. Indeed, recent works \cite{DBLP:conf/interspeech/HollandsBC22} have shown that state-of-the-art ASR models perform significantly worse on non-native speakers than on native speakers when evaluated on datasets with a diverse range of accents, such as the Speech Accent Archive (SAA). However, the SAA benchmark has its limitations as (i) it only contains read speech and (ii) it does not have a good representation of Indian speakers (only 59/1500 speakers).

\begin{figure}[!t]
    \centering
    \includegraphics[width=0.84\linewidth]{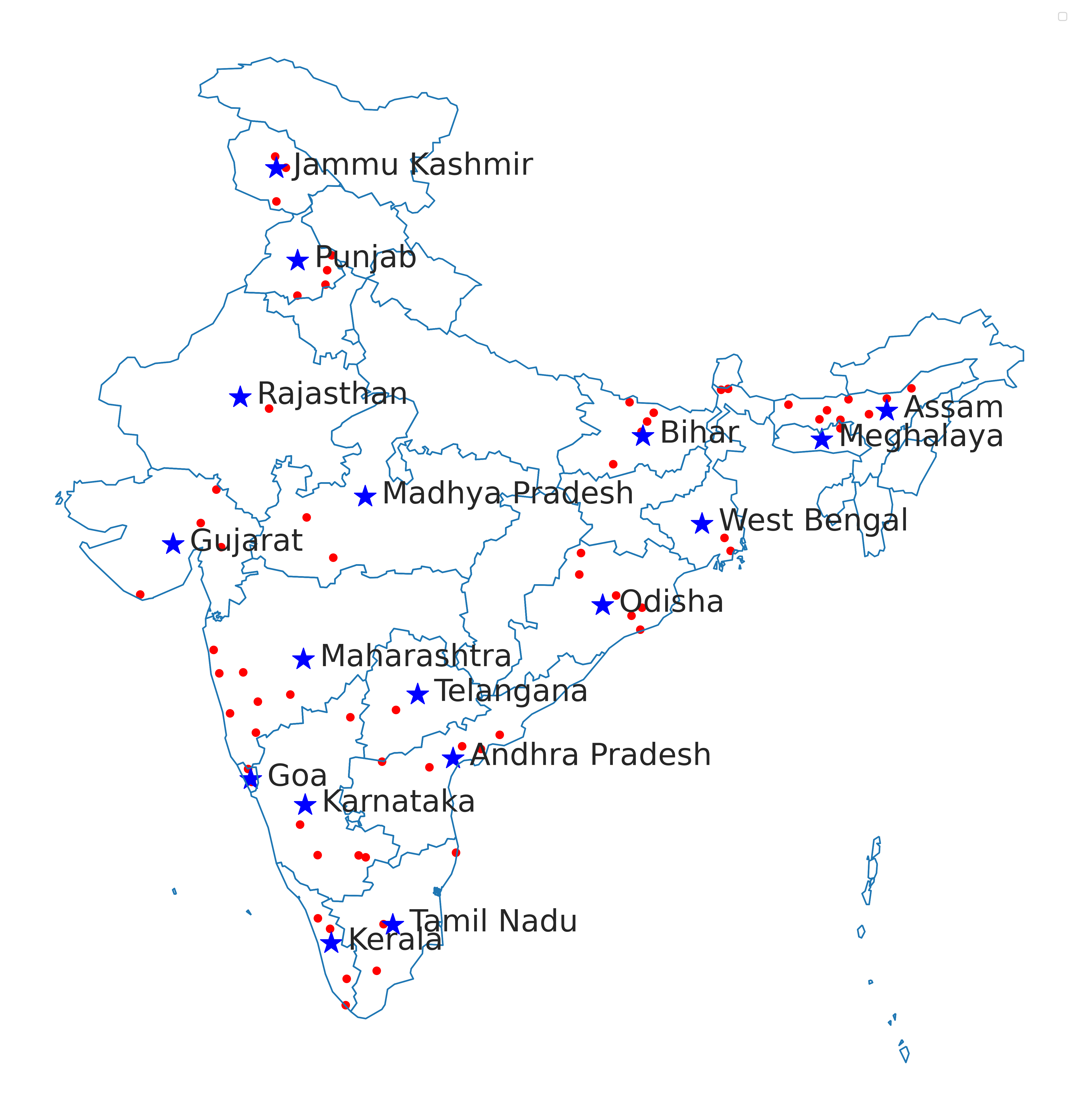}
    \caption{\benchmark~contains data from 117 speakers spanning 65 districts across 19 states in India and speaking 19 out of the 22 constitutionally recognized languages.}
    \label{fig:districts}
\end{figure}

In this work, we focus on creating a robust benchmark for evaluating ASR systems for Indian accents. India is recognized for its rich linguistic diversity, with 22 official languages, 122 major languages, and 1599 other languages in the country, as per the Census of 2011\footnote{https://censusindia.gov.in/nada/index.php/catalog/42561}. Despite the abundance of regional languages, English continues to be a language of importance in India, serving as the official language of the Indian government, the Supreme Court of India, and the primary medium of instruction in higher education institutions. Current estimates show that approximately 10\% or about 130 million people in India speak English, making it the second largest English-speaking country in the world\footnote{https://en.wikipedia.org/wiki/List\_of\_countries\_by\_English-speaking\_population}. In addition, the accents of English speakers in India vary significantly due to the influence of regional languages. For instance, native speakers of Dravidian languages, such as Tamil and Malayalam, have distinct intonation patterns and pronunciation styles compared to native speakers of Indo-Aryan languages, such as Hindi, Marathi, Gujarati, etc. Even within the Indo-Aryan family, there are significant variations in the accents of speakers from the country's northern, western, and eastern parts. Hence, a robust English ASR benchmark covering the accent diversity of India is needed.

To address this gap, we introduce \benchmark, an ASR benchmark containing Indian-accent English data. To build \benchmark, we recruited speakers from diverse geographical locations spanning the length and breadth of India. Figure \ref{fig:districts} shows the distribution of speakers across different parts of the country covering 65 districts across 19 states in India. The benchmark contains a total of 9.6 hours of transcribed data collected from 117 speakers having diverse language backgrounds resulting in different accents. More specifically, the collective set of native languages spoken by the speakers covers 19 of the 22 constitutionally recognized languages of India, belonging to 4 different language families. The dataset includes both read speech and spontaneous conversational data, covering a variety of domains such as history, culture, tourism, government, sports, etc. It also contains data corresponding to popular use cases such as ordering groceries, making digital payments, and using government services (e.g., checking pension claims, checking passport status, etc.). The resulting diversity in vocabulary as well as use cases allows a more robust evaluation of ASR systems for real-world applications. 

Using \benchmark, we evaluate 6 state-of-the-art English ASR models based on different paradigms. These include (i) End-to-end models such as OpenAI's Whisper \cite{whisper} which is trained on weakly supervised web-scale data (438K hours) curated from YouTube which is noisy but very diverse (ii) Self-supervised models such as Wav2Vec2, HuBERT, WavLM, and Data2Vec which use a large amount of unlabeled data for pre-training and relatively smaller amounts of clean supervised data (960 hours) (iii) Conformer based model \cite{DBLP:conf/interspeech/GulatiQCPZYHWZW20} released by NVIDIA which is trained on a large amount of manually labeled data (24.5K hours). We also evaluate two commercial models offered by Google and Microsoft. We observe that the Whisper model which is trained on large amounts of diverse weakly supervised data performs significantly better than the other open-source models mentioned above. Further, two of the six open-source models perform significantly better than both the commercial models and exhibit very little standard deviation across different Indian accents. The benchmark as well as our code will be released publicly and will help in a more robust evaluation of English ASR systems on a diverse set of Indian accents. 

\section{Related Work}
There are several datasets available for training and evaluating English ASR systems. A few popular ones which are freely available under permissible licenses include Librispeech \cite{7178964}, Switchboard-1 Dataset \cite{Godfrey1992SWITCHBOARDTS}, WSJ-0 and WSJ-1 \cite{Drude2019SMSWSJDP}, VCTK \cite{Yamagishi2019CSTRVC}, VoxPopuli \cite{wang-etal-2021-voxpopuli} and Mozilla Common Voice \cite{Ardila2019CommonVA}. However, none of these have a good representation of Indian speakers with diverse accents. For example, Librispeech, Switchboard, and the two WSJ datasets have no speakers residing in India. Similarly, VoxPopuli mainly contains European Parliament event recordings and is unlikely to have a representation of a diverse set of Indian speakers. 
There are a few datasets that specifically address the problem of evaluating ASR systems on multiple accents. These include the Speech Accent Archive (SAA) dataset and The Foreign Accented English (FAE) dataset \cite{SP2/K7EQTE_2022}. SAA only has 59 Indian speakers contributing a total of 26 minutes of data. Similarly, FAE  only consists of utterances spoken by native Hindi and Tamil speakers 
contributing a total of 0.5 hours and 1 hour of data respectively, and does not capture the large linguistic diversity of India. There are three datasets, viz., the IITM and NPTEL datasets\footnote{\label{note1}https://sites.google.com/view/englishasrchallenge/home} and the AccentDB dataset \cite{ahamad-anand-bhargava:2020:LREC} which specifically focus on Indian accent English. However, the IITM and NPTEL datasets contain data only from the STEM domain and the speakers have good fluency in English given that they come from highly educated backgrounds (professors at premier higher education institutes in India). The AccentDB dataset only contains \textit{8} speakers belonging to \textit{4} Indian languages. In contrast to these benchmarks, \benchmark~contains more diverse content from a variety of domains and use cases. Further, it is collected from a diverse set of speakers across India speaking 19 different native languages.

\section{\benchmark}
We now describe the various steps involved in creating \benchmark. 

\noindent\textbf{Recruitment of diverse speakers.} Through a network of professional translators, we contacted speakers who were fluent in English but were native speakers of a regional Indian language. For each of the 19 languages,  we recruited 3-5 bilingual speakers who spoke English and one of the constitutionally recognized Indian languages resulting in a total of 117 speakers. Of these, 54 were men and 63 were women. We also ensured that we had a roughly equal number of speakers belonging to the following age groups 18-30, 30-45, 45-60, and 60+. The speakers also came from different educational backgrounds (arts, commerce, science) with different levels of education (graduates, post-graduates, PhDs). The task was clearly explained to the speakers, and they were informed that the data was being collected to build and evaluate speech models. Their voice samples were recorded only if the speakers willingly agreed to participate in the task and signed a consent form to this effect. 

\noindent\textbf{Collection of multi-domain text data.} A part of the data included read speech which required participants to read a piece of text shown to them. To ensure that the text being read covered diverse vocabulary, we collected text from Wikipedia belonging to multiple domains. These domains were identified using the ``Category'' information available for Wikipedia articles. In total, we collected 1k sentences from 9 domains, viz. health, entertainment, culture, geography, history, business, news, sports, and tourism. Each participant was asked to read 5 sentences randomly chosen from this collection while ensuring that no two participants got the same sentence. 

\noindent\textbf{Creation of prompts for collecting extempore data.} The participants were asked to fill out a form, where, in addition to meta-data such as age, district of residence, native language, etc. they were also asked to select (i) topics of interest and (ii) specific domains about which they could talk. We considered 28 topics of interest such as painting, cooking, gardening, knitting and stitching, travelling, etc., and the same 9 domains listed above. 
For each topic of interest, we created a few simple questions that any participant could answer, such as, ``What inspired you to take up drawing?'', ``What are the dishes you like to cook and tell us the recipe of your favourite dish?''. Similarly for each domain, we created simple questions which anyone interested in that domain could answer. For example, someone interested in Entertainment should be able to answer the following question ``What is your favorite movie or TV serial and why do you like it\text{?}''.  Each participant was shown four such questions for each of the selected topics of interest and domains and was required to answer these questions using extempore speech.

\noindent\textbf{Creation of prompts for different use cases.} One of the goals was to create a benchmark that also contains utterances that one would encounter in popular everyday use cases. We considered three such use cases, \textit{viz.}, ordering grocery items, making digital payments, and using government services. For each of these use cases, we created prompts such as, ``Add 2kg Tomato, 1 Cadbury Chocolate to my shopping cart'', ``Pay 1000 rupees from my SBI bank number 123456789 to UPI ID 9876543210@paytm'' and so on. Each participant was shown five such  prompts for each of these use cases. 

\noindent\textbf{Creation of a mobile app for data collection.} Given the diverse locations from which we intended to collect data, we needed a tool that allowed us to distribute and monitor work remotely. To do so, we used Microsoft’s open-source crowdsourcing platform called Karya which is available as an android application and has already been used by other teams in the past to collect voice data \cite{iitb_msr, DBLP:journals/corr/abs-2208-11761}. Once a participant fills out the participation form, they can install Karya and log in with their mobile number. On logging in, the participant will see different tasks corresponding to (i) read speech (ii) questions on topics of interest (iii) questions on domains of interest (iv) prompts on everyday use cases described above. Once they enter a task, they can read the prompt and press the record button to start recording their response to the prompt. Once done, the participant can press the ``stop'' button, at which point the app automatically replays the audio recorded by the participant. The participants were instructed to listen to the audio to ensure that (i) there was no background noise and (ii) the recording was clearly audible even at 50\% of the volume of the device.    

\noindent\textbf{Transcription of voice samples.} To transcribe the data we recruited undergraduate and postgraduate students who work on various research projects in our institute and are fluent in English (having done their schooling as well a higher education in English). The transcribers were asked to verify the recorded samples to check for audio quality and to also ensure that the responses were on topic (\textit{i.e.}, the participants were responding correctly to the prompts). The transcribers then segmented the data into logical segments (typically, sentences) and then transcribed these segments using a modified instance of Label Studio. Given the extempore and conversational nature of the data, they were asked to use the same transcription guidelines as used for the Switchboard corpus.

In summary, \benchmark\footnote{\benchmark~ means accent in Sanskrit} contains a total of 9.6 hours of transcribed data from a total of 117 Indian speakers from 19 different languages covering 65 districts in 17 Indian states. 
It contains 1.4, 6.4, and 1.7 hours of (i) read speech, (ii) extempore speech on topics of interest, and (iii) utterances from everyday use cases, respectively. The share of extempore speech is larger as participants tend to give longer responses to questions about their topics of interest as opposed to everyday use cases where the responses are to the point.

\section{Methodology}
We evaluate a total of 8 models on \benchmark. Since \benchmark~only contains data from non-native (L2) speakers, to show the contrast with the performance of native (L1) speakers, we consider the LibriSpeech dataset and the Speech Accent Archive (SAA) dataset. In particular, we consider the data from L1 speakers in SAA and evaluate the models on this subset of the data (SAA-L1). Of the 8 models we considered, 6 models are open-source and are publicly available, whereas 2 models are commercially available, one each from Google and Microsoft. From the open-source category, we chose models based on different paradigms as listed below. We do not use an external language model with any of the models listed below as the idea was to evaluate the ability of the acoustic model to handle diverse accents.  

\noindent\textbf{End-to-End models.} Here, we consider OpenAI's Whisper model which has been trained on 680K hours of weekly supervised multilingual data of which 438K hours of data is in English. One advantage of this model is the huge diversity in the training data which perhaps makes it better suited for handling diverse accents. The downside, of course, is that the training data is not entirely gold-standard but weakly supervised. We consider the base, medium and large models available as a part of OpenAI's official release on HuggingFace.

\noindent\textbf{Self-supervised models.} Here, we consider models such as Wav2Vec2, HuBERT, WavLM, and Data2Vec which are pre-trained on large amounts of unsupervised data and then finetuned on a relatively smaller amount of supervised data. We used the publicly available versions of these models from HuggingFace, which were pretrained on 60k hours of LibriVox data and then finetuned on 960h of Librispeech data except WavLM where the authors used a 100-hour clean subset of Librispeech for finetuning the model.

\noindent\textbf{Conformer based models.} All the models described above are transformer-based models. We also consider the conformer-based model trained with CTC loss as released by NVIDIA. This model does not use any unsupervised pre-training but is trained on a larger amount of gold standard data collated from 11 different sources, resulting in a total of 24,500 hours of training data. We hoped that given the diverse and relatively cleaner and larger training data, this model might perform well on diverse accents.

From the commercially available models, we tested Google\footnote{https://cloud.google.com/speech-to-text} and Azure\footnote{https://azure.microsoft.com/en-us/products/cognitive-services/speech-to-text} using their provided SDKs. In addition to evaluating the international version (en-US) of their ASR models, we also evaluate their India-specific models (en-IN). 

\begingroup
\setlength{\tabcolsep}{2.6pt} 
\begin{table}[]
\centering
\begin{tabular}{lrrrr}
\toprule
&\# Params. &Svarah &SAA\_L1&Libri \\
\midrule
Whisper$_{base}$ &74M &13.6 &2.9& 4.2\\
Whisper$_{medium}$ &769M &8.3 &1.7& 3.1 \\
Whisper$_{large}$ &1550M &7.2 &1.6& 2.7\\
\midrule
Wav2Vec2$_{large}$ &317M &24.9 &3.1& 1.8\\
HuBERT$_{large}$ &316M &25.6 &3.2& 2.0\\
WavLM$_{large}$ &300M &33.7 &9.2& 3.4\\
Data2Vec$_{large}$ &313M &24.5 &2.5& 1.8\\
\midrule
Conformer$_{large}$ &120M &14.6 &1.1& 2.1\\
\midrule
Azure$_{US}$ &- &20.9 &24.2& -\\
Azure$_{IN}$ &- &21.3 &30.1&-\\
Google$_{US}$ &- &30.0 &16.8&-\\
Google$_{IN}$ &- &20.7 &63.7 &-\\
\bottomrule
\end{tabular}
\caption{WERs of different models on (i) \benchmark~that contains data from Indian speakers \& (ii) SAA\_L1, LibriSpeech Clean (Libri) which contain data from native English speakers.}
\label{tab:benchmarkresults}

\end{table}
\endgroup

\section{Results \& Discussions}
The results of our experiments are summarised in Tables \ref{tab:benchmarkresults} and \ref{tab:spliced-wers}. Below, we discuss the main observations from these results.

\begingroup
\setlength{\tabcolsep}{2pt} 
\begin{table}[t]
\centering
\begin{tabular}{p{0.08\linewidth}p{0.12\linewidth}p{0.12\linewidth}p{0.12\linewidth}p{0.10\linewidth}p{0.10\linewidth}p{0.10\linewidth}p{0.10\linewidth}}
\toprule
 &W$_{large}$ &W2V2 & HuB & D2V &C$_{large}$ &G$_{IN}$ & Az$_{US}$   \\
\midrule
\multicolumn{8}{l}{\textbf{Split by accents}} \\
\midrule
ne &9.8 &34.4 &35.2 &32.6 &20.5 &27.5 &33.1 \\
brx &11.6 &29.9 &30.5 &29.0 &18.5 &26.7 &38.3 \\
as &10.1 &32.3 &33.8 &31.6 &21.4 &20.6 &18.1 \\
doi &8.0 &29.1 &28.5 &26.8 &16.5 &21.1 &21.6 \\
sd &7.3 &27.2 &26.5 &26.5 &16.8 &19.1 &21.0 \\
ml &8.1 &23.7 &26.5 &23.0 &14.7 &24.9 &20.4 \\
kn &6.6 &27.2 &27.1 &27.0 &14.6 &17.7 &14.2 \\
mr &6.8 &24.6 &26.0 &26.2 &15.0 &18.6 &15.5 \\
bn &7.6 &24.1 &24.8 &25.1 &14.1 &18.5 &16.6 \\
pa &6.7 &22.3 &22.3 &22.6 &13.3 &17.4 &25.2 \\
ur &6.2 &22.3 &23.3 &21.8 &14.2 &21.5 &19.7 \\
te &7.5 &23.4 &23.1 &22.6 &14.3 &18.8 &17.6 \\
ks &6.9 &20.3 &21.2 &20.3 &12.6 &27.8 &18.3 \\
kok &6.4 &20.1 &21.3 &19.7 &11.9 &19.2 &25.3 \\
or &6.2 &24.5 &24.9 &24.8 &12.1 &17.5 &13.4 \\
guj &6.0 &23.3 &24.0 &22.6 &12.7 &13.1 &17.8 \\
hi &5.3 &20.7 &20.8 &20.4 &10.9 &19.7 &21.0 \\
mai &4.5 &20.7 &21.9 &20.7 &11.6 &15.4 &19.6 \\
ta &5.3 &17.8 &18.8 &18.6 &10.7 &17.4 &9.9 \\
\midrule
\multicolumn{8}{l}{\textbf{Split by type of data}} \\
\midrule
Re &6.2 &22.8 &23.3 &21.7 &12.6 &21.3 &24.0 \\
Ex &7.4 &23.4 &25.3 &28.0 &14.0 &18.4 &10.1 \\
UC &11.2 &34.6 &35.4 &33.7 &23.3 &19.8 &15.7 \\
\bottomrule
\end{tabular}
\caption{The WER of different models on different splits of \benchmark~based on (i) accents and (ii) types of data (Re: read speech, Ex: extempore speech, UC: use cases). Model names are abbreviated as W$_{large}$: Whisper, W2V2: Wav2Vec2, HuB: HuBERT, D2V: Data2Vec, G$_{IN}$: Google$_{IN}$, Az$_{US}$:  Azure$_{US}$.}
\label{tab:spliced-wers}
\end{table}
\endgroup

\begin{figure}[t]
    \centering
    \includegraphics[width=0.7\linewidth]{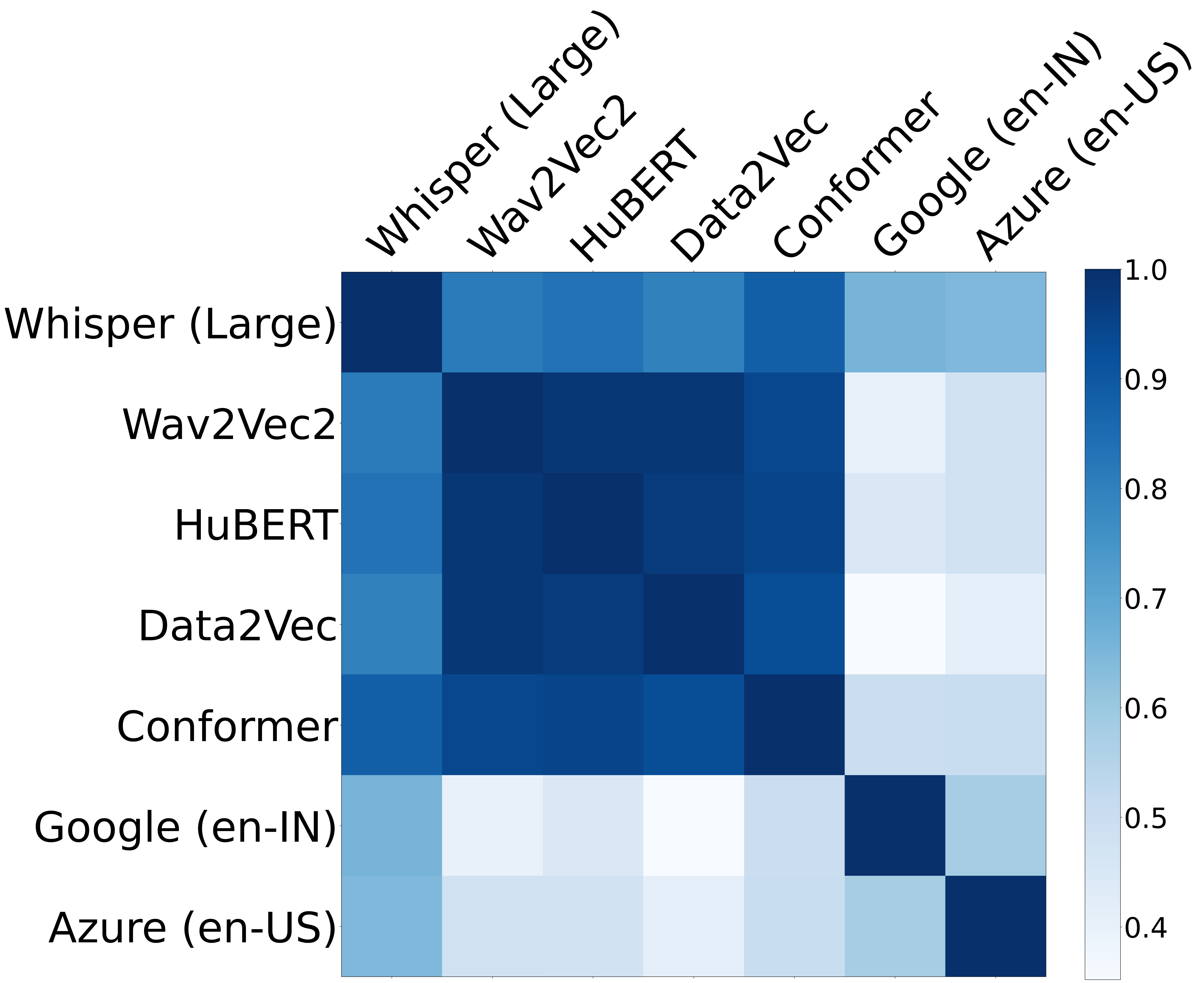}
    \caption{Correlation between the WERs of different models across 
different accents.}
    \label{fig:corr}
\end{figure}

\noindent\textbf{Poor performance on non-native speakers.} Referring to Table \ref{tab:benchmarkresults}, we observe that for all the 6 open source models there is a significant gap between the performance of the models on \benchmark~and L1 data (SAA\_L1 and LibriSpeech). More specifically, the gap between the WERs on \benchmark~ and SAA\_L1 ranges from 5.65 - 43.03. Similarly, the gap between the WERs on \benchmark~ and LibriSpeech ranges from 4.5 - 30.3, clearly indicating that the claims of human parity need to be revisited by evaluating on more diverse benchmarks like \benchmark. Note that the medium and large variants of Whisper give single-digit WERs which is encouraging but, given the large size of these models, it may be difficult to deploy them in the Indian context. Surprisingly, both the Azure models perform better on \benchmark~than on the L1 data. It is hard to comment on why this is the case as the details of Azure's models are not known.

\noindent\textbf{Comparison of different models.} Again referring to Table \ref{tab:benchmarkresults}, we observe that the Whisper family of models gives the best performance on \benchmark~with the smallest gap in WERs on \benchmark~and L1 data. We hypothesize that this is due to the large diversity in the weakly supervised training data used by these models which makes them more robust to diverse accents. In other words, the trade-off between the quality and volume of the training data works favorably. The next best-performing model is the NVIDIA Conformer-based model which was again trained on a large amount of gold standard data collated from different sources (again reinforcing the importance of diversity in training data). The self-supervised models (Wav2Vec2, HuBERT, WavLM and Data2Vec) perform poorly, indicating that pre-training on large amounts of diverse unlabelled data is not as useful as finetuning on diverse data. WavLM which is trained on the smallest amount of training data (100 hours) gives the worst performance. Lastly, the two commercial models are outperformed by the open-source Whisper and NVIDIA Conformer models, making a clear case for using open-source models. 

\noindent\textbf{Performance across different accents.} Table \ref{tab:spliced-wers} shows the performance of the models split by the native language spoken by the speakers. Due to space constraints, we consider only the top-7 models from Table \ref{tab:benchmarkresults}. 
We observe that the best-performing models, viz., Whisper-Large and NVIDIA Conformer, have very low standard deviation across different accents (1.8 and 3.0 respectively). 
In general, the models perform poorly for speakers who are native speakers of Assamese, Bodo, and Nepali which happen to be low-resource languages from the North-Eastern region of India. Figure \ref{fig:corr} shows that there is a high correlation between the performance of the open-source models across accents, indicating that in general, the relative performance of models across different accents is similar. Further, the correlation between the 3 self-supervised models is higher given the similar pre-training and fine-tuning data used by these models. Both the commercial models do not correlate well with any of the open source models perhaps due to different training data (including proprietary data) and design choices.

\noindent\textbf{Performance across different types of data.} As mentioned earlier, \benchmark~contains read speech, extempore data and utterances corresponding to everyday use-cases. Referring to the last section of Table \ref{tab:spliced-wers}, we observe that except for the two commercial models, all models perform better on read and extempore data as compared to the data corresponding to use cases. This is mainly  because the data corresponding to everyday use cases contains many entities such as brand names, bank names, food items, document IDs, and so on. Recognizing such entities when spoken in Indian accents is hard for existing ASR systems. Even the Whisper$_{large}$ model performs poorly on the utterances from everyday use cases with a WER of 11.2, indicating that there is much scope for improvement.

\section{Conclusion}
In this work, we addressed the lack of Indian-accented English data in existing speech recognition benchmarks by introducing \benchmark, a dataset containing 9.6 hours of transcribed data collected from 117 speakers across India, covering a broad range of accents and linguistic backgrounds. Our dataset includes both read speech and spontaneous conversational data, covering a variety of domains such as history, culture, tourism, government, sports, etc. We evaluated six open-source and two commercial ASR models on this dataset and found that the recently released Whisper model outperformed commercial models by a significant margin. Our dataset and all evaluation scripts will be publicly available. The creation of \benchmark~will enable a more robust evaluation of ASR systems for real-world applications in India, where English is an important language, and a diverse range of accents exists due to the influence of regional languages.

\section{Acknowledgements}
We would like to thank the Ministry of Electronics and Information Technology (MeitY\footnote{https://www.meity.gov.in/}) of the Government of India and the Centre for Development of Advanced Computing (C-DAC\footnote{https://www.cdac.in/index.aspx?id=pune}), Pune for generously supporting this work and providing us access to multiple GPU nodes on the Param Siddhi Supercomputer. We would like to thank the EkStep Foundation and Nilekani Philanthropies for their generous grant which went into hiring human resources as well as cloud resources needed for this work. We would like to thank the AI4Bharat team for helping us to connect to native speakers of different languages across the country. 

\bibliographystyle{IEEEtran}
\bibliography{mybib}

\end{document}